\pdfoutput=1

\documentclass[11pt]{article}

\usepackage{EMNLP2022}

\usepackage{times}
\usepackage{latexsym}

\usepackage[T1]{fontenc}

\usepackage[utf8]{inputenc}

\usepackage{microtype}

\usepackage{inconsolata}
\usepackage{microtype}
\usepackage{listings}
\usepackage[most]{tcolorbox}
\usepackage{color}
\definecolor{dkgreen}{rgb}{0,0.6,0}
\definecolor{gray}{rgb}{0.5,0.5,0.5}
\definecolor{mauve}{rgb}{0.58,0,0.82}

\usepackage{microtype}
\usepackage{booktabs}
\usepackage{multirow}
\usepackage{geometry}
\usepackage{pdflscape}
\usepackage{rotating}
\usepackage{bbm}

\definecolor{codepink}{rgb}{0.98, 0.25, 0.3}
\definecolor{codeblue}{rgb}{0.00, 0.53, 0.99}
\definecolor{codegreen}{rgb}{0.38,0.1,0.82}
\definecolor{myblue}{rgb}{0.9, 0.1, 0.94}
\definecolor{mygreen}{rgb}{0.64, 0.56, 0.88}
\definecolor{myyellow}{rgb}{0.98, 0.94, 0.75}
\definecolor{mygreen}{rgb}{0.68, 0.9, 0.6}
\definecolor{mydarkgreen}{rgb}{0.0, 0.8, 0.6}
\definecolor{myred}{rgb}{255, 0, 0}
\definecolor{myorange}{rgb}{255, 165, 0}

\usepackage{enumerate}
\newenvironment{itemize*}%
 {\leftmargini=10pt\begin{itemize}%
  \setlength{\itemsep}{0pt}%
  \setlength{\parskip}{0pt}%
  }%
 {\end{itemize}}
\newenvironment{enumerate*}%
 {\begin{enumerate}%
  \setlength{\itemsep}{0pt}%
  \setlength{\parskip}{0pt}}%
 {\end{enumerate}}

\title{KGxBoard: Explainable and Interactive Leaderboard for Evaluation of \\ Knowledge Graph Completion Models}

\author{Haris Widjaja\thanks{\hspace{0.15cm}Equal contribution}\hspace{0.15cm}$^1$, Kiril Gashteovski\footnotemark[1]\hspace{0.15cm}$^2$, Wiem Ben Rim$^2$, Pengfei Liu$^{1,4}$, \\ 
\textbf{Christopher Malon$^3$, Daniel Ruffinelli$^5$, Carolin Lawrence$^2$, Graham Neubig$^{1,4}$}  \\
  $^1$ Carnegie Mellon University; $^2$ NEC Laboratories Europe; \\
  $^3$ NEC Laboratories America; $^4$ Inspired Cognition; $^5$ University of Mannheim;}

\begin{document}
\maketitle
\begin{abstract}
Knowledge Graphs (KGs) store information in the form of \emph{(head, predicate, tail)-}triples. 
To augment KGs with new knowledge, researchers proposed models for KG Completion (KGC) tasks such as link prediction; i.e., answering \emph{(h; p; ?)} or \emph{(?; p; t)} queries. Such models are usually evaluated with averaged metrics on a held-out test set. 
While useful for tracking progress, 
averaged single-score metrics cannot reveal \emph{what exactly} a model has learned---or failed to learn. 
To address this issue, we propose KGxBoard\footnote{All resources (code, data, demo, video, etc.) are available on \url{https://github.com/neulab/KGxBoard}}: an interactive framework for performing fine-grained evaluation on meaningful subsets of the data, each of which tests individual and interpretable capabilities of a KGC model.  
In our experiments, we highlight the findings that we discovered with the use of KGxBoard, which would 
have been impossible to detect with standard averaged single-score metrics.
\end{abstract}

\section{Introduction}

Knowledge Graphs (KGs) are graph databases that store information about entities and the relations between them in the form of \emph{(head, predicate, tail)-}triples \cite{weikum2021machine}. Because of their flexible structure, KGs are used for storing general real-world data 
\cite{rebele2016yago} as well as domain-specific data covering 
various domains \cite{abu2021domain}, including medicine \cite{chandak2022building}, IoT \cite{le2016graph} and  finance \cite{cheng2020knowledge}. 
KGs play an important role in NLP: they are used for many downstream tasks, including  
language modeling \cite{RobertLLogan2019BaracksWH}, entity linking \cite{radhakrishnan2018elden} and question answering \cite{Saxena2022SequencetoSequenceKG}.


\begin{figure*}
    \centering
    \includegraphics[width=0.95\textwidth]{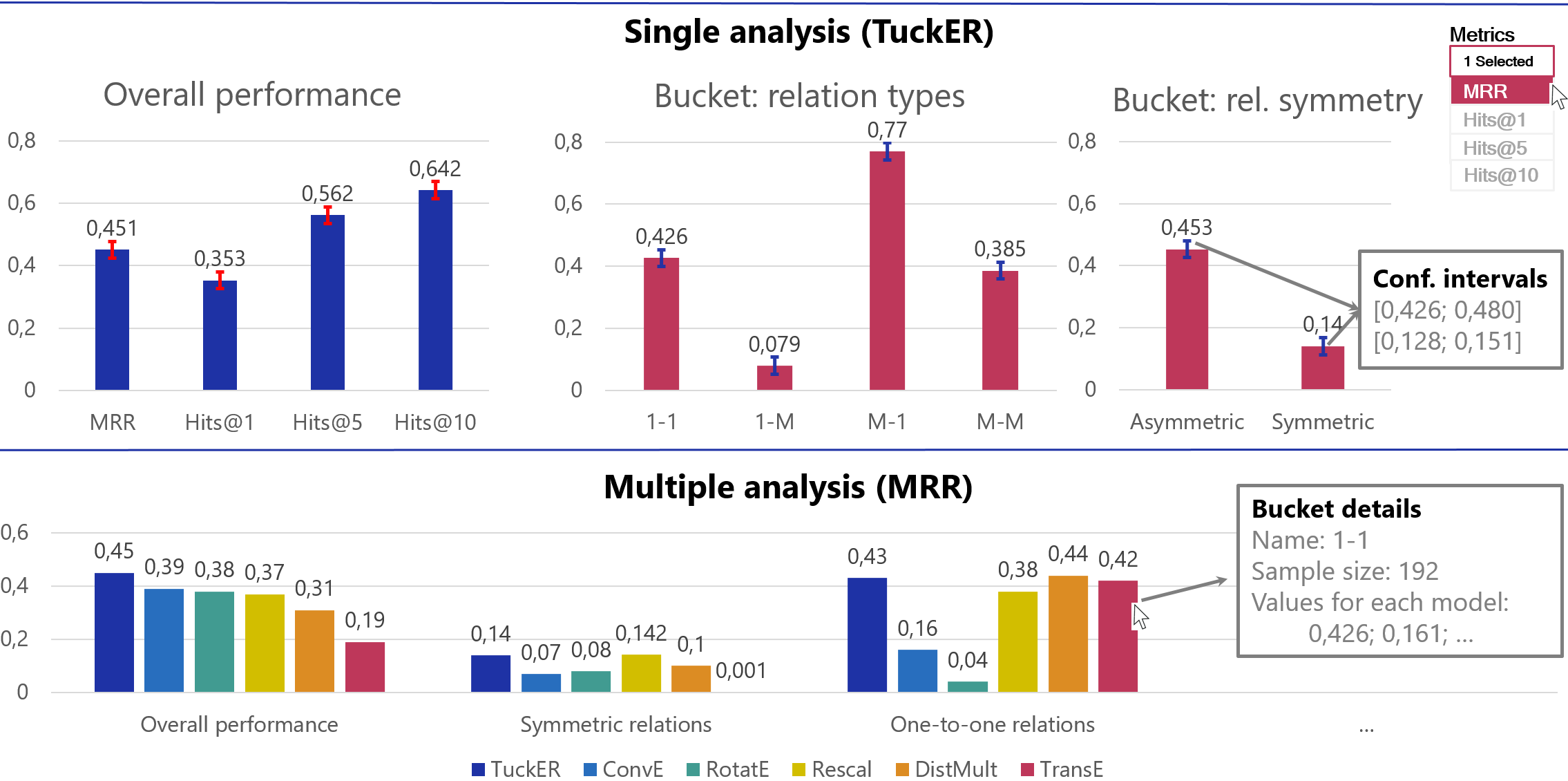}
    \caption{Illustration of KGxBoard's functionalities. With the \emph{single analysis mode}, the user can see the overall performance of a given KGC model, as well as the performance of the model across multiple buckets (i.e., partitions of the evaluated data). For example, we observed that the TuckER model performs very well on triples with Many-to-one (M-1) relations. However, it performs poorly on triples with One-to-many (1-M) relations and symmetric relations. With the \emph{multiple analysis mode}, the user can compare the overall and bucketized performance across different models. Such a view enables the user to compare the models in a fine-grained manner; e.g., Rescal is ranked 4th in the overall performance, but it is ranked 1st for the triples having symmetric relations. Likewise, DistMult is ranked 5th in the overall performance and 1st for the triples with One-to-one relations. In both modes, the user can interact with the UI to see details of the data buckets, change the evaluation metrics, etc.}
    \label{fig:my_label}
\end{figure*}

A common problem for KGs is that they are incomplete (i.e., they do not contain all facts about the world or the particular domain), which could lead to limitations in performance for the downstream tasks. To tackle this problem of incompleteness, the research community has worked on many KG Completion (KGC) tasks, most prominently on \emph{link prediction}:
predicting new facts within the KG by providing ranked predictions to the queries \emph{(h; p; ?)} or \emph{(?; p; t)} \cite{clouatre2021mlmlm}. 
Models for KGC tasks are typically evaluated with 
single-score metrics that are averaged over a held-out 
test set. 
For instance, for a KGC query---e.g., \emph{(h; p; ?)}---, hits@k indicates the average number of correct answers from the test set that appear within the top-k ranked entities predicted by the KGC model.\footnote{See Sec.~\ref{sec:prelims} for more details about hits@k and other metrics}

While using such scores is important for tracking the progress of KGC models, researchers observed that a more fine-grained evaluation is needed \cite{palmonari2020knowledge}, because the averaged metrics cannot answer the question of \emph{what properties} have the models actually learned---or failed to learn. To investigate what properties were learned by the KGC models, 
researchers have designed specific datasets and experimental setups. For example, \citet{rim2021behavioral} used 
the idea of behavioral testing applied to NLP models \cite{ribeiro2020beyond} to perform more fine-grained tests for relation symmetry. 
In particular, they measured the performance of KGC models for queries that contain symmetric relations; i.e., relations that are true for both \emph{(h; p; t)} and \emph{(t; p; h)} such as \emph{(X; marriedTo; Y)}.
However, their proposed framework only contains a limited number of tests and might not cover model properties of interest to other researchers. 

To make the fine-grained evaluation more generic---and to compare different KGC models across different properties---we propose KGxBoard: a method and software implementation for fine-grained evaluation of KGC models (see illustration in Fig.~\ref{fig:my_label}). 
KGxBoard splits the evaluated data into groups (buckets) according to certain properties of the data. For instance, one can split the test data into two buckets of data points such that 
one bucket contains triples with symmetric and the other with asymmetric relations. Consequently, the users can observe the performance of each of these buckets separately, while also having an overview of the overall performance scores. 

KGxBoard builds upon ExplainaBoard \cite{liu2021explainaboard}, which is an explainable leaderboard for NLP tasks. We adapt ExplainaBoard to the KGC tasks\footnote{KGxBoard can handle the KGC tasks on answering queries of the form \emph{(?, p, t), (h, ?, t)} and \emph{(h, p, ?)}. For simplicity, when we refer to KGC models in this paper, we refer to models that provide predictions of the form \emph{(h, p, ?)}.} by providing:
(1) a method and software implementation for fine-grained evaluation of KGC models, integrated into ExplainaBoard; 
(2) APIs for porting results from two popular KGC frameworks---PyKeen \cite{ali2021pykeen} and LibKGE \cite{broscheit2020libkge}---directly into KGxBoard format; 
(3) interface for reading customized features for fine-grained evaluation; (4) experimental study exposing problems with KGC models that cannot be spotted with averaged scoring; (5) experimental study showing that the findings from the fine-grained evaluation of KGxBoard can be used for automatic debugging of the models. 



\section{Preliminaries}
\label{sec:prelims}


The performance of KGC models is evaluated on a held-out test set of golden KG triples. In particular, the models return a set of ranked answers to the queries---e.g., \emph{(h; p; ?)}---, where the correctness of the answers is evaluated against the golden triples. The final score is averaged over the examples from the held-out test set. The standard scores used in the literature are:

\begin{itemize}
    \item[(1)] $\text{Hits}@k= \frac{1}{|\mathcal{D}|} \sum_{(h, r, t)\in \mathcal{D}} \mathbbm{1} [\mathtt{rank}(t) \leq k] $

\item[(2)] $\text{MRR} = \frac{1}{|\mathcal{D}|} \sum_{(h, r, t)\in \mathcal{D}} \frac{1}{\mathtt{rank}(t)}$

\item[(3)] $\text{MR} = \frac{1}{|\mathcal{D}|} \sum_{(h, r, t)\in \mathcal{D}} \mathtt{rank}(t)$

\end{itemize}
\noindent where $\mathcal{D}$ is the set of test triples, and $h, r, t$ are the head, relation and tail of the KG triple respectively. 

\section{KGxBoard's Fine-grained Evaluation}
Prior work in NLP has pointed to the issues of using single-score metrics, which do not expose what exact properties of the data were (or were not) learned by the models \cite{narayan2021personalized}; e.g., some information extraction models perform poorly when there is a conjunction present in the sentence \cite{gashteovski2022benchie}. To better understand what properties were (or were not) learned by the NLP models, people have proposed multifaceted or explainable and interactive leaderboards (for more details on related work, see Appendix~\ref{app:rel_work}).

Following ExplainaBoard \cite{liu2021explainaboard}, the basic idea of KGxBoard's fine-grained evaluation 
is to breakdown the performance measure (e.g., Hits@10) over individual groups (buckets) in addition to the performance score over the overall evaluation dataset. This approach involves three steps: (i) define features upon which the evaluation dataset is going to be partitioned; (ii) partition evaluation dataset into different buckets based on the defined features; and (iii) calculate performance w.r.t.~each bucket. In contrast to ExplainaBoard, KGxBoard is tailored for the KG completion tasks and their evaluation metrics.


\subsection{Feature definition}

The feature definition describes the manner upon which KGxBoard is going to partition the evaluated data into buckets. For example, if the feature is about predicate symmetry, then the data is divided in 2 buckets: (1) triples that have predicates which are considered symmetric;\footnote{predicates that are true for both
\emph{(h; p; t)} and \emph{(t; p; h)}; e.g., the predicate \emph{p=marriedTo}: \emph{(x; p; y)}$\iff$\emph{(y; p; x)}.} (2) triples with asymmetric predicates. Because each data point can be assigned to one of these buckets with a label \emph{"symmetric"} or \emph{"asymmetric"} (strings), the user defines the feature "predicate symmetry" accordingly (as shown on Figure~\ref{fig:feature_def}).

\begin{figure}
\begin{lstlisting}
"custom_features": {
    "rel_type": {  
        "dtype": "string",
        "description": "predicate symmetry",
        "num_buckets": 2
    }
}
\end{lstlisting}
\caption{Defining custom features for bucketization of the evaluation data (e.g., validation or test data).}
\label{fig:feature_def}
\end{figure}

\paragraph{Built-in Features.}
KGxBoard supports several built-in features that will automatically bucketize the evaluation data of any models to be analyzed. The built-in features include several widely used properties of the data, such as predicate symmetry \cite{trouillon2016complex} and entity type hierarchy \cite{rim2021behavioral}; 
see details in Appendix~\ref{app:built-in-features}.


\paragraph{Customized Features.}
KGxBoard also allows users to customly define their own features by specifying additional information in the system output file. If, for example, the users want to define the bucketization features for predicate symmetry, then they only need to specify this in a json file (Fig.~\ref{fig:feature_def}).

\subsection{Partitioning of Evaluation Data into Different Buckets}

For the built-in features, KGxBoard automatically assigns each data point to its respective bucket. For the customized features, once the custom features were defined, the user should place each data point to its respective customized bucket via the bucketization functions (explained in Section~\ref{subsec:bucketizatoin-functions}). This data is then fed into KGxBoard, which automatically computes the relevant metrics for each bucket.

\subsection{Calculate Performance w.r.t.~each Bucket}

KGxBoard computes the relevant metrics (described in Section~\ref{sec:prelims}) for each bucket individually, as well as for the overall evaluation dataset. 

\paragraph{Confidence Interval.}
KGxBoard has been endowed with the ability to quantify to what degree we can trust the result of each bucket. Specifically, as illustrated in Figure~\ref{fig:my_label}, each bin has been equipped with an error bar and its width reflects the reliability of the bucket performance. KGxBoard supports two ways to calculate the confidence interval: bootstrapped re-sampling~\cite{efron1992bootstrap} and t-test~\cite{nakagawa2007effect} .


\section{KGxBoard: System Overview}


\begin{figure}
    \includegraphics[width=\linewidth]{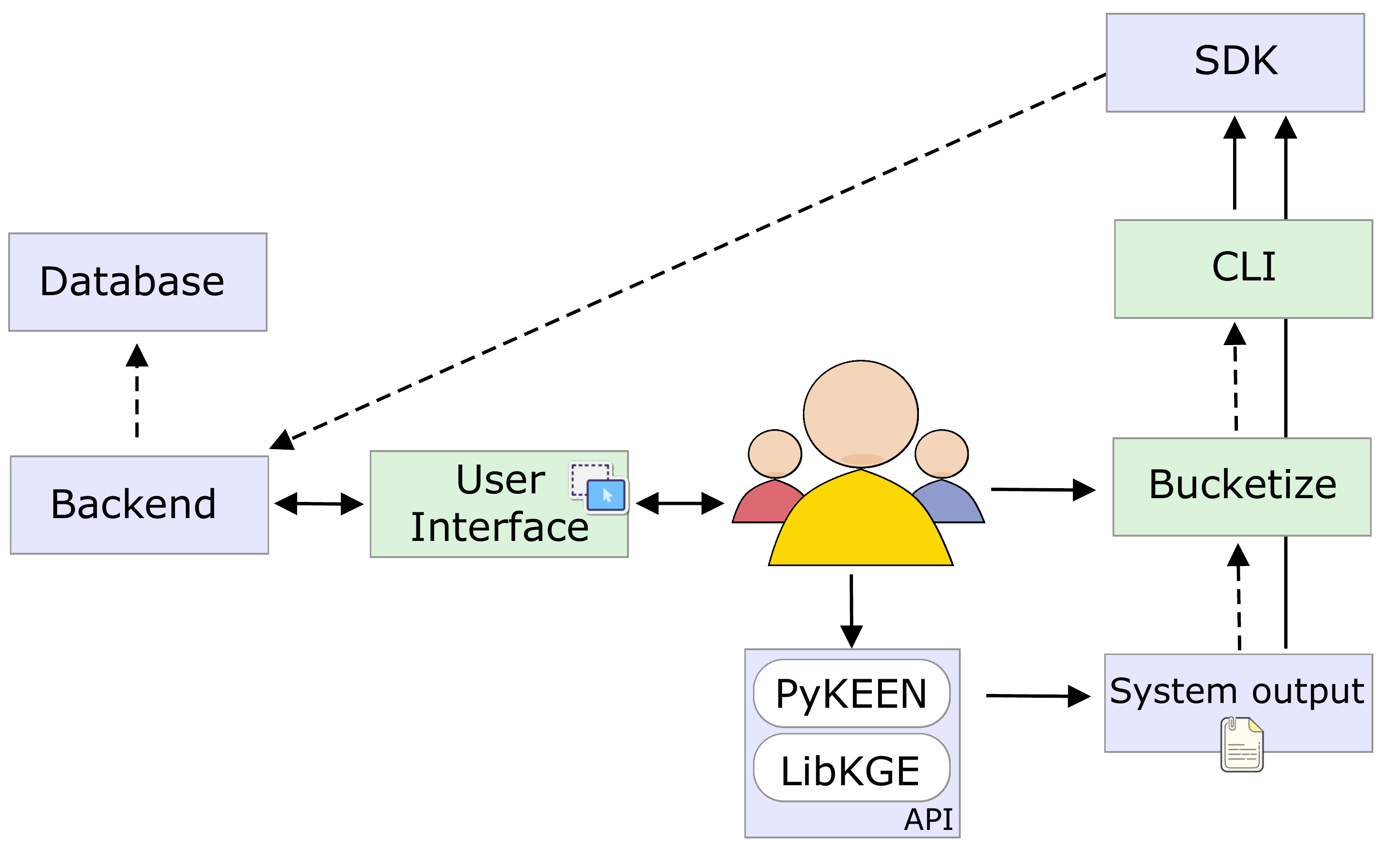}
  \caption{General overview of KGxBoard's architecture. SDK: Software Development Kit; CLI: Command-Line interface.}
  \label{fig:overview}
\end{figure}

A general overview of the KGxBoard architecture is illustrated in Figure~\ref{fig:overview}. The users can provide the data for the models through the front-end (via the UI) or the back-end (via the CLI or the SDK). Then, these results are stored in a database (DB), which can be accessed and viewed either with the visual interface, programmatically, or with the command-line interface. In general, the users can choose three ways to use KGxBoard's functionalities: (i) directly from the interactive web interface; (ii) through an API from KGC frameworks (PyKEEN and LibKGE); (iii) through the command-line interface with already provided data. 

\subsection{Frontend}

\begin{figure}
  \includegraphics[width=\linewidth]{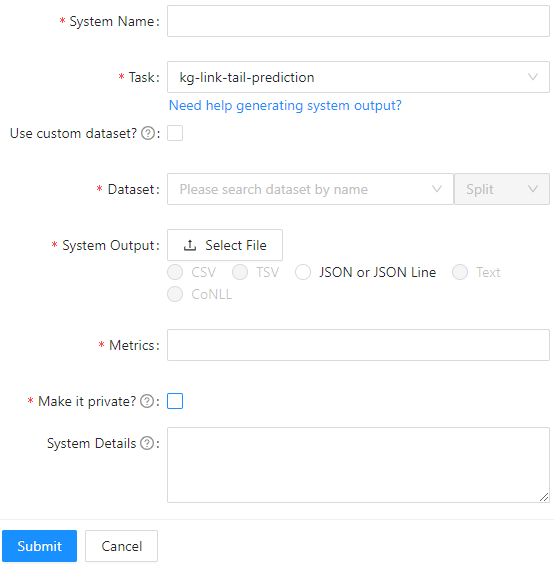}
  \caption{Using the frontend to upload a new model.}
  \label{fig:frontend-data-upload}
\end{figure}

We adopt a React-based technology stack\footnote{\url{https://reactjs.org/}} to create an interactive web app as the frontend of KGxBoard. 
Assuming that a user has generated the input data in prior steps (e.g., through the API from KGC libraries; see Section~\ref{subsec:api-kge}), she can upload the data via the frontend interface (Figure~\ref{fig:frontend-data-upload}). The data is then passed on to the backend, which stores it in a database. 
The frontend provides two types of analysis of the models' evaluation:
\textbf{single analysis} aims to identify the strengths and weaknesses of a given KG completion model;
\textbf{pairwise/multiple analysis} can help users figure out where one model is better (or worse) then the other when two (or more) models are selected; for illustration of the frontend, see Figure~\ref{fig:my_label}.

\subsection{Backend}
\label{subsec:backend}

The backend is built on top of ExplainaBoard's backend code \cite{liu2021explainaboard}. The main functionalities for KGxBoard's backend are: (i) defining the evaluation metrics for the link prediction task: Hits@k, MRR and MR; (ii) computing the overall scores and the buckets' scores with the built-in features; (iii) handling customized feature buckets if the user provided any; (iv) storing the results in the DB; (v) communicate the results with the frontend in an interactive manner.

\subsection{API with KGC Libraries}
\label{subsec:api-kge}

KGxBoard comes with APIs that can translate the output of two widely used KGC libraries---PyKEEN \cite{ali2021pykeen} and LibKGE \cite{broscheit2020libkge}---into KGxBoard format. In particular, the APIs write the system-output files in KGxBoard format which does not contain buckets, only results from the models. If the user wishes to add customized buckets, she can either write or reuse already existing bucketization function(s), which will rewrite the data in KGxBoard format with the desired bucketizations.

\subsection{Bucketization functions}
\label{subsec:bucketizatoin-functions}

The bucketization functions are procedures that do two main actions: (1) define the buckets; (2) assign a bucket label to each data example. Here's an example of a bucketization function pseudocode that defines relations as being either symmetric or assymetric:

\begin{lstlisting}
# s_rels -> set of symmetric rels.
def bucketize_rel_sym(s_rels):
    # define the buckets properties
    bucket_name = "rel_sym"; dtype = "string"
    descr="rel's symmetry prop."; num_buckets=2
    
    # assign bucket to example data
    for triple in predict_data:
        if triple['predicate'] in s_rels:
            triple.bucket = 'symmetric'
        else:
            triple.bucket = 'asymmetric'
\end{lstlisting}

\section{Experimental Study}


To showcase the usefulness of KGxBoard, we conducted the following experimental study: (1) bucketized comparison of models: we provide insights on the buckets' performance about different models trained with one KGC framework, which would have been impossible to discover with standard metrics; (2) comparison of models trained on different KGC frameworks with different hyperparameter settings: showing how KGxBoard can be used to discover differences between models trained in different environments; (3) automatic debugging: showing the ability of automatic debugging of the models by using insights from certain buckets.

\subsection{Experimental Setup}

\paragraph{Datasets.} We used two widely used datasets: (1) FB15K-237 \cite{toutanova2015observed},  constructed from Freebase \cite{Bollacker2008FreebaseAC}, covering relations between people, locations, etc.; (2) WN18RR \cite{Dettmers2018Convolutional2K}, constructed from WordNet \cite{Miller1992WordNetAL} and represents connections between words in English, such as synonyms and hypernyms. The models were trained, validated and tested with the standard dataset split (the reported results  are on the test sets). 

\paragraph{Models and KGC Frameworks.} 
For our experiments, we trained the KGC models on two commonly used KGC frameworks: PyKEEN and LibKGE. With PyKEEN we trained the following models for link prediction: DistMult \cite{yang2014embedding}, ConvE \cite{Dettmers2018Convolutional2K}, RESCAL \cite{Nickel2011ATM}, RotatE \cite{sun2019rotate}, TransE \cite{bordes2013translating} and TuckER \cite{Balazevic2019TuckERTF}. 
PyKEEN provides hyperparameters to reproduce the results of the original work of a given model, which we used to train the models. To compare the same models trained with different hyperparameters and frameworks, we also used the pretrained models obtained by \citet{ruffinelli2020you} as a result of extensive hyperparameter optimization using LibKGE \cite{broscheit2020libkge}. Specifically, we used the models ConvE, DistMult, Rescal and TransE (see training details in App.~\ref{app:pykeen_training_details}).

\paragraph{Bucketizations.} We partitioned the test data into different interpretable groups based on either built-in or customized features (e.g., relation type). 
All together, for FB15K-237 / WN18RR we have 303 / 35 unique buckets respectively. The large difference in the number of buckets between the datasets is mainly due to the significantly higher number of relations in the FB15K-237 dataset (237 v.s.~11). We provide further details on the buckets in App.~\ref{app:buckets_list}.


\subsection{Bucketized Comparison of Models}
\label{subsec:bucketized_exp}

\begin{table}
    \centering
    \footnotesize
        \begin{tabular}{rccc}  
            
            \toprule
            
                    & Overall rank  & $b^{=}$   & $b^{\neq}$      \\ \midrule
            TuckER  &   1 / 1         &  .601 / .543   &   .399 / .457       \\ 
            ConvE   &   2 / 3         &  .301 / .486   &   .699 / .514      \\ 
            RotatE  &   3 / 5         &  .257 / .686   &   .743 / .314      \\ 
            Rescal  &   4 / 2         &  .261 / .457   &   .739 / .543    \\ 
            DistMult&   5 / 4         &  .541 / .400   &   .459 / .600     \\ 
            TransE  &   6 / 6         &  .868 / .800   &   .132 / .200     \\ 
            \bottomrule
        \end{tabular}
    
    \caption{Ranking of KGC models (trained with PyKEEN) according to the MRR score. 1 indicates best rank. $b^{=}/b^{\neq}$: the fraction of buckets where the ranking of a given model is equivalent/not equivalent compared to the overall rank of the model. Results are on the datasets FB15K-237 / WN18RR.} 
    \label{tab:ranking_buckets}
\end{table}

To get an overview of the fine-grained evaluation of the KGC models, we used the predictions from PyKEEN on the FB15K-237 and WN18RR datasets. By ranking different systems with the MRR metric in two ways: (1) based on their overall performance; and (2) based on bucket-wise performance; we obtain that, as shown in Table~\ref{tab:ranking_buckets}, the overall ranking of the models is significantly different than the ranking of the models for each individual bucket. 

For example, TuckER is ranked as the best-performing model on both FB15K-237 and WN18RR. However, TuckER is not ranked as best performing model for approximately 40\% and 46\% of the buckets for FB15K-237 and WN18RR respectively. For instance, when taking a closer look at FB15K-237, we find that for the bucket featurized by symmetric relations, TuckER is ranked 2nd and Rescal is ranked 1st (on the overall test set, Rescal is ranked 4th) and for the One-to-one relations, TuckER is also ranked 2nd and DistMult is ranked 1st (on the overall test set, DistMult is ranked 5th); see Figure~\ref{fig:my_label}.

Such findings would be impossible to spot with standard averaged metrics over the entire test set, and are similar in spirit to previous results that show how alternative evaluation methods can expose differences in overall model performance~\cite{wang2018evaluating,rim2021behavioral}. With KGxBoard, researchers can diagnose issues with any KGC model on customized properties of the data (i.e., customized bucketizations of the evaluation data). 


\subsection{Bucketized Comparison of Models Trained with Different Hyperparameters}
\label{subsec:orighp_vs_optimhp}

Similarly as with the fine-grained comparison between diffferent models (Sec.~\ref{subsec:bucketized_exp}), KGxBoard can be used to compare one model that was trained on multiple sets of different hyperparameters (HPs) and implementations. For this purpose, we train each of the four KGC models (ConvE, Rescal, DistMult and TransE) with two sets of HPs on two KGC libraries---PyKEEN and LibKGE---and showcase the differences. Note that the implementation from different libraries can vary significantly w.r.t.~the HP search space and the degree of customization.\footnote{For instance, while LibKGE allows the customized initialization of the embeddings through HPs, PyKEEN requires changing the source code. More details in App.~\ref{app:pykeen_training_details}.}



On the one hand, we trained each model with the hyperparameters that aim to reproduce the work of the original papers (trained with PyKEEN; the HP combination was proposed in PyKEEN's documentation). On the other hand, we used model configurations resulting from a HP optimization pipeline that ensures improved overall results (trained with LibKGE; HP combination proposed by \citet{ruffinelli2020you}). 
We refer to each of the former models as \textsc{OrigHP-Model} and to the latter as \textsc{OptimHP-Model}.  While the \textsc{OptimHP} models always outperform their \textsc{OrigHP} counterparts, we still observed many cases where the ranking of the models flipped on some buckets. For example, \textsc{OrigHP-ConvE} performs better than \textsc{OptimHP-ConvE} for: (1) the FB15K-237 triples with relation between award ceremony and award winner; (2) FB15K-237 triples whose tail entity types are of type "musical work" (details in App.~\ref{app:apis_comparison}).

\subsection{Automatic Debugging of Models Using Buckets} 


\begin{table}[htb]
\footnotesize
\begin{center}
\begin{tabular}{ccccc}
\hline
Hits@1 & ConvE & TuckER & RotatE & Rescal \\
\hline
\multicolumn{5}{c}{\em Debugging test} \\
Before debug. & .0000 & .0000 & .0000 & .0000 \\
Naive & .0625 & .1875 & .0465 & .1642 \\
In-danger & .1562 & .2083 & .0465 & .2015  \\ \hline
\multicolumn{5}{c}{\em Original test} \\
Before debug. & .2710 & .3108 & .2627 & .2596 \\
Naive & .2416 & .3010 & .2627 & .2594 \\
In-danger & .2574 & .3004 & .2627 & .2594  \\
\hline
\end{tabular}
\caption{Debugging results for the relations with most symmetry violations (Hits@1).}
\label{tbl:debugging}
\end{center}
\end{table}


Grouping related properties of the model into buckets offers not only the
potential to diagnose problems, but also the potential to fix them by
debugging.
We show here how the debugging techniques of \citet{Malon2022FastFD}
may be adapted to fix problems with KG embeddings,
illustrating the idea with one particular bucket: relation symmetry.

Updating the relation embedding to improve symmetry
for one relation will have no effect on other relations generally,
so we debug only one relation $r$ at a time.
For debugging, we collect a sub-bucket consisting of triples $(h,r,t)$ that
violate symmetry in a stricter sense: the reverse triple $(t,r,h)$ is
in the training set and the trained model predicts $h$ as the tail
for $(t,r,?)$ with rank one, but $t$ has rank greater than one
among the predictions for $(h,r,?)$.
Ten triples from the sub-bucket constitute the {\em debugging set},
which is used to learn better model parameters, and the remaining triples
are held out to form the {\em debugging test set}.

We debug the relation with the most symmetry violations (in the above sense)
for four models: ConvE, TuckER, RotatE and Rescal.\footnote{The other two models, Distmult
and TransE, did not have enough symmetry violations to fill a debugging set.}
A naive method, which we call {\em intensive fine-tuning},
simply fine-tunes the model on the debugging set alone until all
its triples are predicted at rank one.  We freeze entity embeddings during
intensive fine-tuning, because updating the entity embedding will not
generalize to improve symmetry on any held-out entities.
To monitor whether this learning
comes at the expense of forgetting other triples, we evaluate performance
on the original test set before and after debugging, in addition
to the debugging test set.

We also adapt the proposed method of \citet{Malon2022FastFD} to KGE.
In this method, we first run the intensive fine-tuning, then collect
twenty triples at random from the training set, which were correctly
predicted at rank one after the original rank-tuning, but whose rank fell
after the intensive fine-tuning. We use these twenty examples
together with the ten debugging examples in a second round of
intensive fine-tuning, again starting with the parameters of the original
model.  The hope is to learn the debugging examples while anchoring
the performance of triples that are ``in danger'' of being forgotten.

Table~\ref{tbl:debugging} shows debugging results for the four models.
In all cases, naive debugging improves Hits@1 on the held-out test
debugging examples, and in-danger debugging often yields a further
improvement.  The impact of debugging on Hits@1 of the original
test set is less than 1\% for all models except ConvE, which
has interaction parameters that are applied to many relations.
For ConvE, the in-danger method cuts this sacrifice in half.
We provide more detailed analysis in Appendix~\ref{sec:debuggingfull}.


\section{Conclusions}

We presented KGxBoard: interactive framework for fine-grained and interpretable evaluation on meaningful subsets of the data, each of which tests individual and interpretable capabilities of a KGC model. 
We highlighted insights that would be impossible to detect with standard leaderboards. 


\clearpage

\bibliography{anthology,custom}
\bibliographystyle{acl_natbib}

\clearpage

\section*{Appendix}

\appendix



\section{Discussion on Related Work}
\label{app:rel_work}


\subsection{Evaluation of NLP Models}

Recent work in NLP has pointed to the problems of using single-score metrics \cite{ethayarajh2020utility,narayan2021personalized,Bowman2021WhatWI,kotnis2022human}. In particular, such single-score metric evaluations do not expose the data-centric properties that were not learned by the models (e.g., problems with learning named entities that span across multiple tokens). In NLP, such issues are tackled by proposing multifaceted and/or explainable leaderboards and benchmarks \cite{liu2021explainaboard,gashteovski2022benchie,Friedrich2022AnnIEAA,kotnis2022milie,xiao2022datalab,Kiela2021DynabenchRB,Thrush2022DynataskAF}. To the best of our knowledge, there is no prior work that proposes such multi-faceted leaderboards for KG completion models. 

\subsection{Evaluation of KGC Models}


\paragraph{KGC Benchmarks.} Prior work has shown that learning latent representations---i.e., KG Embeddings (KGEs)---for the entities and relations is highly beneficial for tackling KGC tasks \cite{link_pred_efficient,ruffinelli2020you}. Researchers have evaluates the effects of hyperparameters (and different training strategies) of KGE models for link prediction. In particular, \citet{kadlec2017knowledge} studied the effects of different training objectives on the DistMult model \cite{yang2014embedding} and found that using the cross entropy loss function is a better alternative to the binary cross entropy.  \citet{ruffinelli2020you} performed a systematic study on many KGE models with the LibKGE framework \cite{broscheit2020libkge} and found that training components make a huge difference in model performance. \citet{safavi2020codex} proposed a more data-centric benchmark (CoDEx), which improves upon previous benchmarks by proposing additional datasets extracted from Wikidata \cite{vrandevcic2014wikidata} and then using them to improve current KGE models. \citet{sun2020re} proposed an evaluation framework such that it breaks ties of same-score answers according to different strategies. PyKEEN \cite{ali2021pykeen} is another KGE framework and benchmark, which facilitates the training and evaluation of KGE models for KG completion tasks across a variety of datasets. Although highly useful for overall evaluation, none of these frameworks provide fine-grained evaluation of KGE models, nor an interactive interface for such evaluation.

\paragraph{Metrics.} Other line of work focused on proposing different metrics for exposing different problems within the KGE models \cite{wang2018evaluating}, though these metric are averaged scores over the entire test set and do not  provide additional insights about where the models make mistakes. Prior work has shown that the standard averaged evaluation metrics hits@k and MRR favor popular entities and relations \cite{mohamed2020popularity}. Contrary to such approaches, KGxBoard operates across the standard metrics (hits@k, MRR and MR), but in addition it supports highly customized fine-grained analysis and interactive interface.

\paragraph{Studying Specific Properties.} Other line of work targets specific properties of the models, i.e., evaluating if a set of KGE models learned some specific properties. For example, relation symmetry is a property that has been extensively studied in the literature \cite{trouillon2016complex,sun2019rotate,peng2020lineare,zhang2020learning,wang2014knowledge}. Other line of work  investigated other properties, such as entity type hierarchy \cite{rim2021behavioral,zhang2020learning}, (inverse) equivalence \cite{meilicke2018fine}, subsumption \cite{meilicke2018fine}, relation and entity frequency \cite{mohamed2020popularity} and entity distribution \cite{bordes2013translating}. 

\paragraph{KG Analyzers.} Recently, there were proposals for systems that analyse already constructed KGs \cite{ilievski2020kgtk}. Such work, however, does not cover the evaluation of KGC models, it only covers analysis of already existing KGs.



\section{Details on Built-in Features}
\label{app:built-in-features}

KGxBoard comes with built-in features that bucketize the data automatically. Some of the built-in features in KGxBoard are dataset-specific, while others are dataset-agnostic.
 
\begin{itemize*}
    \item Length of head/tail entity: the number of tokens in the head/tail entity (dataset-agnostic).
    \item Frequency of head/tail entity: the frequency of tail entity in the training set (dataset-agnostic).
    \item Frequency of the predicate: the frequency of the predicate in the training set (dataset-agnostic).
    \item Symmetry of relation: the symmetry of entity relations (dataset-specific; for now KGxBoard supports FB15K-237).
    \item Entity type level: the most specific (highest) entity type level of true tail entity (dataset-specific, for now we support FB15K-237). In particular, we mapped each Freebase entity to its DBpedia \cite{auer2007dbpedia} counterpart, and then used DBpedia's type information \cite{Gangemi2012AutomaticTO} in order to determine the most specific type level that is available in the data; e.g., if we have information \emph{(Barack Obama; type; Person)} and \emph{(Barack Obama; type; Politician)}, we use the latter because it is more specific.
\end{itemize*}


\section{Details on Customized Buckets}
\label{app:buckets_list}

To provide more fine-grained analysis that goes beyond the built-in features for bucketization, as well as to showcase the ability of KGxBoard to handle customized buckets, we partition the test data into buckets, based on the following customized features:

\paragraph{Buckets by relations.} Each data point is placed in a bucket according to its relation. For example, the triples \emph{(England; /location/location/contains; Lancaster)} and \emph{(Los Angeles; /location/location/contains; Beverly Hills)} are placed in the same bucket because they have the same relation. This bucketization is dataset-agnostic and we used it for both FB15K-237 and WN18RR.

\paragraph{Buckets by relation types 1-1, 1-M, M-1 or M-M.} Following \citet{bordes2013translating}, we partition the data into four possible buckets according to their relation properties: 1-to-1 (1-1), 1-to-many (1-M), many-to-1 (M-1) and many-to-many (M-M). According to this definition, each relation has a property on how many entities it can have as a head or tail. For example, the relation \emph{isAuthorOfBook} is 1-M relation (because one author can be the author of several books) and the relation \emph{sportsTeamLocation} is 1-1 relation (because one sports team can have only one home location). 

\paragraph{Tail entity type (level 1).} The entity type (level 1, as described by \citet{rim2021behavioral}) of the gold entity that needs to be predicted. This customized feature is specific for FB15K-237 and not for WN18RR. In particular, we leverage similar approach as with the built-in entity type level feature (described in Appendix \ref{app:built-in-features}): we mapped each Freebase entity to its DBpeda counter part and then used DBpedia's type information to determine the type at level 1 of the entity type hierarchy.

\paragraph{Tail entity type (level 2).} The entity type (level 2, as described by \citet{rim2021behavioral}) of the gold entity that needs to be predicted. As with the previous customized feature, this customized feature is specific for FB15K-237 and not for WN18RR. 

\paragraph{Relation's symmetry (for WN18RR).} In Appendix \ref{app:built-in-features}, we described KGxBoard's built-in feature that bucketizes the data into symmetric and asymmetric relations for the FB15K-237 dataset. This feature, however, is not natively supported by KGxBoard for the WN18RR dataset. For WN18RR, we followed \citet{rim2021behavioral} and (1) got a set of all unique relations from WN18RR; (2) manually defined which relation is symmetric and which one is not. Then, we assigned a bucket (symmetric or asymmetric) to each data point in the test set.

\section{Training Details for PyKEEN and LibKGE}
\label{app:pykeen_training_details}

PyKEEN provides hyperparameters to reproduce the results of the original work of a given model, which we used to train RotatE, ConvE, TuckER, TransE and Rescal on both \textsc{WN18RR} and \textsc{FB15K237} datasets. After training TransE, the hyperparameters for \textsc{WN18RR} returned a very low MRR result indicating that it failed to train for the task, and for DistMult, no configuration files corresponding to the above-mentioned datasets were provided. In such cases, we used insights from LibKGE \cite{ruffinelli2020you} to train the models with PyKEEN to the best of our abilities; to implement the same models in the two frameworks, we match some parameters such as the number of epochs, embedding dimensions, initializer function, optimizer and learning rate arguments, including the scheduler and its parameters. However, the hyperparameters of a KGE model on the mentioned frameworks are not matched one-to-one. LibKGE allows the user to set both the initializer of the relation and entity embeddings along with setting the lookup embedder weight, patience, regularizer, dropout. In contrast, PyKEEN only allows the naming of the initializer function. Moreover, while negative sampling is possible in both frameworks, PyKEEN allows setting the negative sampler function that describes how to generate corrupt triples for training and allows the setting of negatives per positive triples as well as the filtering and corruption scheme, where LibKGE allows the previous as well as adding the number of samples for each head, relation, object. We expect that these differences in implementation and hyperparameters will have an impact on the results seen on KGcBoard. We provide the used hyperparameters and trained models with our code.

It is worth noting that PyKEEN does provide configurations for the same models with optimized parameters, but since the MRR and Hits@k results of these models are outperformed by the LibKGE models, we chose to make this comparison to both feature the difference in hyperparameters as well as the framework implementation in our experiments. For the models trained on LibKGE, we used the pretrained models obtained by \citet{ruffinelli2020you} as a result of hyperparameter optimization using LibKGE \cite{broscheit2020libkge}.

\section{Detailed Results for Bucketized Comparison of Models}
\label{app:apis_comparison}

In these experiments, we showcase how the users can use KGxBoard in order to compare one model trained on multiple hyperparameter settings. As explained in Section \ref{subsec:orighp_vs_optimhp}, on the one hand, we trained each model with the hyperparameters that aim to reproduce the work of the original papers (dubbed \textsc{OrigHP-Model}); on the other hand we used pretrained models that use a hyperparameter optimization pipeline that ensures improved overall results (dubbed \textsc{OptimHP-Model}). Each model trained with the \textsc{OrigHP} hyperparameter settings was trained with PyKEEN, by using the hyperparameters combination that aims to reproduce the results from the original papers and was proposed in PyKEEN's documentation. Each model trained with the \textsc{OptimHP} hyperparameter settings was trained with with LibKGE, by using the hyperparameter combination proposed by \citet{ruffinelli2020you}. The results are summarized in Table~\ref{tab:ranking_buckets_apis2}.

 While the \textsc{OptimHP} models always outperform their \textsc{OrigHP} counterparts in both FB15K-237 and WN18RR datasets, we still observed many cases where the ranking of the models flipped on some buckets. For example, \textsc{OptimHP-ConvE} on FB15K-237 performs better than \textsc{OrigHP-ConvE} on the overall score, as well as on 88\% of all the buckets. However, \textsc{OrigHP-ConvE} performs better than \textsc{OptimHP-ConvE} for several buckets, including (1) the FB15K-237 triples with relation between award ceremony and award winner; (2) FB15K-237 triples whose tail entity types are of type "musical work". 
 In a more extreme case for FB15K-237, when we compared \textsc{OptimHP-TransE} with \textsc{OrigHP-Transe}, the optimized \textsc{OptimHP-TransE} is ranked as 1st on the overall score and in 98\% of the buckets (this is to be expected, given that the difference in the overall score is 23 percentage points). Yet, in 2\% of the buckets (4 buckets in total), \textsc{OrigHP-TransE} is ranked as 1st. 

\begin{table}
    \centering
    \footnotesize
        \begin{tabular}{rcccc}  
            
            \toprule
            
            Model & Overall rank  & MRR score & $b^{=}$   & $b^{\neq}$      \\ 
                     & \multicolumn{2}{c}{(\textsc{OptimHP} / \textsc{OrigHP})}              &                & \\ \midrule
                    \multicolumn{5}{c}{\emph{FB15K-237}} \\ 
            ConvE   &   1 / 2         & 0.44 / 0.39    &  0.88     & 0.12     \\ 
            Rescal  &   1 / 2        & 0.45 / 0.38    &  0.83     & 0.17     \\ 
            DistMult&   1 / 2       &  0.44 / 0.31   &  0.91     & 0.09     \\ 
            TransE  &   1 / 2       &  0.42 / 0.19  &  0.98      & 0.02     \\ \midrule
                \multicolumn{5}{c}{\emph{WN18RR}} \\ 
            ConvE   &   1 / 2       &  0.46 / 0.28            &    1.00        &  0.00 \\
            Rescal   &   1 / 2       &  0.48 / 0.29            &  1.00          & 0.00 \\
            DistMult   &   1 / 2       &    0.48 / 0.27          & 1.00           & 0.00 \\
            TransE   &   1 / 2       &      0.24 / 0.13        &  0.91          & 0.09 \\
            \bottomrule
        \end{tabular}
    
    \caption{Comparison of KGC models trained with different hyperparameter settings. The shown results compare each model between its \textsc{OptimHP} and \textsc{OrigHP} hyperparameter settings. The models trained with \textsc{OptimHP} are trained with hyperparameter optimization pipeline that ensures improved overall results. The models trained with \textsc{OrigHP} use hyperparamter settings that aim to replicate the results from the original papers.  
    1 in \emph{Overall rank} indicates better rank. The models were trained and tested on the FB15K-237 and WN18RR datasets. $b^{=}/b^{\neq}$ indicates the fraction of buckets where the overall rank is equivalent/not equivalent as the bucket's rank.} 
    \label{tab:ranking_buckets_apis2}
\end{table}

\section{Detailed Results for Debugging}
\label{sec:debuggingfull}

\begin{table*}[htb]
\small
\begin{center}
\begin{tabular}{p{1.1in}p{.8in}p{.8in}p{.8in}p{.8in}p{.8in}}
\hline
Model / relation & Hits@1 & Hits@5 & Hits@10 & MR & MRR \\
\hline
{\em ConvE/dated} & & & & & \\
\multicolumn{6}{c}{\em Debugging test} \\
Before debugging & .0000 & .9062 & 1.0000 & 3.3125 & .3556 \\
Naive & .0625 & .3125 & .5312 & 11.7188 & .1863 \\
In-danger & .1562 & .7812 & .9688 & 3.9688 & .3943 \\
\multicolumn{6}{c}{\em Original test} \\
Before debugging & .2710 & .4734 & .5690 & 186.0528 & .3677 \\
Naive & .2416 & .4464 & .5433 & 237.9980 & .3402 \\
In-danger & .2574 & .4614 & .5595 & 203.3516 & .3546 \\
\hline
{\em Tucker/adjoins} & & & & & \\
\multicolumn{6}{c} {\em Debugging test} \\
Before debugging & .0000 & .8542 & .9583 & 3.7708 & .3895 \\
Naive & .1875 & .8542 & .9167 & 5.0208 & .4749 \\
In-danger & .2083 & .8333 & .9167 & 6.7917 & .4678 \\
\multicolumn{6}{c}{\em Original test} \\
Before debugging & .3108 & .5396 & .6283 & 106.4038 & .4171 \\
Naive & .3010 & .5262 & .6162 & 126.6328 & .4055 \\
In-danger & .3004 & .5232 & .6140 & 119.0497 & .4041 \\
\hline
{\em RotatE/friendship} & & & & & \\
\multicolumn{6}{c}{\em Debugging test} \\
Before debugging & .0000 & .0000 & .0000 & 41.0465 & .0419 \\
Naive & .0465 & .1860 & .4302 & 18.9651 & .1493 \\
In-danger & .0465 & .1860 & .4535 & 21.2791 & .1468 \\
\multicolumn{6}{c}{\em Original test} \\
Before debugging & .2627 & .4605 & .5432 & 346.0619 & .3554 \\
Naive & .2627 & .4605 & .5432 & 345.7814 & .3554 \\
In-danger & .2627 & .4605 & .5432 & 345.8078 & .3554 \\
\hline
{\em Rescal/adjoins} & & & & & \\
\multicolumn{6}{c}{\em Debugging test} \\
Before debugging & .0000 & .7761 & .9030 & 5.5597 & .3385 \\
Naive & .1642 & .7313 & .8507 & 9.2687 & .4038 \\
In-danger & .2015 & .6716 & .8060 & 9.1194 & .4040 \\
\multicolumn{6}{c}{\em Original test} \\
Before debugging & .2596 & .4665 & .5562 & 295.5532 & .3574 \\
Naive & .2594 & .4665 & .5560 & 295.6199 & .3573 \\
In-danger & .2594 & .4665 & .5560 & 295.7509 & .3573 \\
\hline
\end{tabular}
\caption{Debugging results for the relations with most symmetry violations.}
\label{tbl:debuggingfull}
\end{center}
\end{table*}

Table~\ref{tbl:debuggingfull} shows debugging results for ConvE,
Tucker, RotatE, and Rescal.
In all cases, naive debugging improves Hits@1 on the held-out test
debugging examples, and in-danger debugging often yields a further
improvement.  In cases where debugging Hits@5 and Hits@10 were high to
begin with, debugging sometimes worsens these metrics, because most
of the debugging examples will be teaching the model to swap the order
of examples already in the top 5 or 10, rather than bring something new
into the top 5 or 10 hits.

Only on ConvE does the naive method reduce original Hits@1 by more than 1\%.
This impact is possible because ConvE (and Tucker) have interaction parameters
that can affect other relations, where RotatE and Rescal have only
relation and entity embeddings, so that the debugging affects only the
relation being debugged.  On ConvE, the in-danger method reduces original Hits@1,
5, and 10 by about half as much as the naive method, while significantly
improving all metrics on the debugging set.  For the other models,
the naive method can achieve simple and effective debugging.



\end{document}